\documentclass[runningheads]{llncs}

\usepackage[year=2026]{eccv}

\usepackage{eccvabbrv}

\usepackage{graphicx}
\usepackage{booktabs}
\usepackage{multirow}
\usepackage{xcolor}
\usepackage{colortbl}
\usepackage{enumitem}
\usepackage{subcaption}
\usepackage{comment}
\usepackage{arydshln}
\usepackage{tikz}
\usetikzlibrary{arrows.meta, positioning, fit, backgrounds, calc, decorations.pathreplacing}

\usepackage[accsupp]{axessibility}

\usepackage{xr-hyper}

\usepackage[breaklinks,colorlinks,citecolor=eccvblue]{hyperref}

\definecolor{darkgreen}{rgb}{0.0, 0.5, 0.0}
\definecolor{lightgray}{gray}{0.9}
\definecolor{bestcell}{rgb}{0.9, 0.95, 1.0}

\newcommand{\best}[1]{\textbf{#1}}
\newcommand{\second}[1]{\underline{#1}}

\newcommand{\loss}[1]{{\scriptsize\textcolor{red}{(#1)}}}

\title{MoCHA: Denoising Caption Supervision for Motion-Text Retrieval}
\titlerunning{MoCHA: Denoising Caption Supervision for Motion-Text Retrieval}

\author{
Nikolai Warner\inst{1}
\and Cameron Ethan Taylor\inst{1}
\and Irfan Essa\inst{1}
\and Apaar Sadhwani\inst{2}
}

\authorrunning{Warner et al.}

\institute{
Georgia Institute of Technology, Atlanta, GA, USA\\
\email{\{nwarner30,irfan\}@gatech.edu}
\and
Amazon, Sunnyvale, CA, USA\\
\email{apaars@gmail.com}
}

\begin{document}

\maketitle

%==============================================================================
\begin{abstract}
%==============================================================================

Text-motion retrieval systems learn shared embedding spaces from motion-caption pairs via contrastive objectives. However, each caption is not a deterministic label but a sample from a distribution of valid descriptions: different annotators produce different text for the same motion, mixing motion-recoverable semantics (action type, body parts, directionality) with annotator-specific style and inferred context that cannot be determined from 3D joint coordinates alone. Standard contrastive training treats each caption as the single positive target, overlooking this distributional structure and inducing within-motion embedding variance that weakens alignment.
We propose MoCHA, a text canonicalization framework that reduces this variance by projecting each caption onto its motion-recoverable content prior to encoding, producing tighter positive clusters and better-separated embeddings. Canonicalization is a general principle: even deterministic rule-based methods (e.g., stopword stripping) improve cross-dataset transfer, though learned canonicalizers provide substantially larger gains. We present two learned variants: an LLM-based approach (GPT-5.2) and a distilled FlanT5 model requiring no LLM at inference time. MoCHA operates as a preprocessing step compatible with any retrieval architecture.
Applied to MoPa (MotionPatches), MoCHA sets a new state of the art on both HumanML3D (H) and KIT-ML (K): the LLM variant achieves 13.9\% T2M R@1 on H (+3.1pp) and 24.3\% on K (+10.3pp), while the LLM-free T5 variant achieves +2.5pp / +8.1pp. Canonicalization reduces within-motion text-embedding variance by 11--19\% and improves cross-dataset transfer substantially---H$\to$K by +94\% and K$\to$H by +52\%---demonstrating that standardizing the language space yields more transferable motion-language representations.

\end{abstract}

%==============================================================================
\section{Introduction}
\label{sec:intro}
%==============================================================================

\begin{figure}[t]
\centering
\footnotesize
% Skeleton strip
\includegraphics[width=0.92\textwidth]{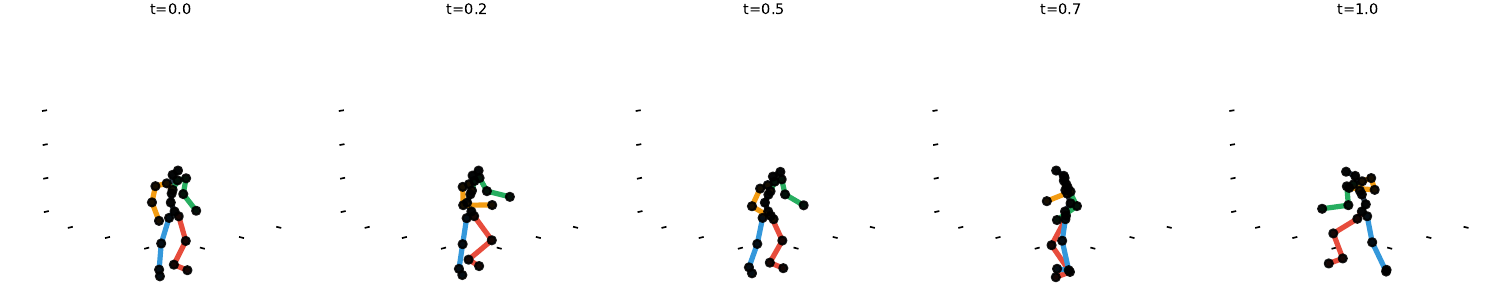}
\vspace{-2pt}
\begin{tikzpicture}[
  capbox/.style={draw=black!40, rounded corners=2pt, inner sep=3pt, font=\scriptsize},
  canonbox/.style={draw=blue!40, rounded corners=2pt, inner sep=4pt, font=\scriptsize, fill=blue!5},
  arr/.style={-{Stealth[length=3pt]}, thick},
]

% --- Left: noisy captions ---
\node[capbox, fill=red!4, text width=10.5cm] (c1) at (0,0) {%
  ``a person \textcolor{red!60!black}{nervously} \textcolor{blue!70!black}{walks forward}, \textcolor{blue!70!black}{stops}, \textcolor{blue!70!black}{turns around}, and \textcolor{blue!70!black}{walks back}.''
  \hfill\textcolor{gray}{\tiny ann.~1}};
\node[capbox, fill=red!4, text width=10.5cm, below=2pt of c1] (c2) {%
  ``a person is \textcolor{blue!70!black}{walking around} and \textcolor{red!60!black}{they are scared}.''
  \hfill\textcolor{gray}{\tiny ann.~2}};
\node[capbox, fill=red!4, text width=10.5cm, below=2pt of c2] (c3) {%
  ``a person \textcolor{blue!70!black}{walks forward and back} \textcolor{red!60!black}{glancing frighteningly} \textcolor{red!60!black}{to his sides}.''
  \hfill\textcolor{gray}{\tiny ann.~3}};

% Label
\node[above=2pt of c1, font=\scriptsize\bfseries, text=black!70] {Same motion, three annotators:};

% Arrow + canonical
\node[canonbox, text width=10.5cm, below=6pt of c3] (canon) {%
  \textbf{MoCHA:}~~``\textcolor{blue!70!black}{walk forward $\to$ stop $\to$ turn around $\to$ walk back}''
  \hfill\textcolor{gray}{\tiny deterministic, motion-recoverable}};

\draw[arr, blue!50!black] (c3.south) -- (canon.north);

% Legend
\node[font=\tiny, text width=10.5cm, below=4pt of canon.south west, anchor=north west] {%
  \textcolor{blue!70!black}{$\blacksquare$~motion-recoverable semantics ($s$)}\quad
  \textcolor{red!60!black}{$\blacksquare$~nuisance factors ($a$): stylistic variation}};

\end{tikzpicture}
\caption{\textbf{Each caption is a different sample from a distribution of valid descriptions.} Three annotators describe the same motion (top) with different captions, each mixing motion-recoverable semantics~$s$ (\textcolor{blue!70!black}{blue}) with annotator-specific nuisance factors~$a$ (\textcolor{red!60!black}{red})---stylistic variation. Standard contrastive training treats each as the single correct target; MoCHA projects each onto~$s$, producing a single deterministic positive.}
\label{fig:motivation}
\end{figure}
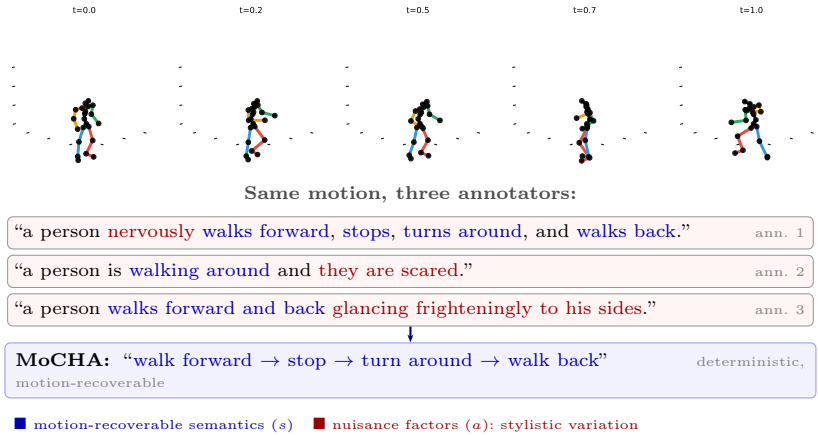

Motion-text retrieval---the task of matching natural language descriptions with 3D human motion sequences---is a foundational capability for motion generation~\cite{petrovich2023tmr,guo2022humanml3d}, action understanding, and human-computer interaction. Recent contrastive approaches~\cite{petrovich2023tmr,bensabath2024crossdataset,motionpatches} learn joint embedding spaces by treating each motion-caption pair as clean ground truth. We argue that this treatment is misspecified: caption supervision is structurally noisy, with each annotation mixing kinematically grounded content with annotator style, hallucinated intent, and non-kinematic additions that have little bearing on the joint coordinates. This noise structure has not been formally analyzed in the motion retrieval literature.

Multi-caption datasets expose this as a many-to-one problem: the same motion of walking forward, stopping, and turning back receives ``nervously walks forward, stops, turns around, and walks back,'' ``walking around and they are scared,'' and ``walks forward and back glancing frighteningly''---three independent draws from a distribution over valid descriptions (Fig.~\ref{fig:motivation}). The shared motion content is buried under annotator-specific style and speculation: ``nervously,'' ``scared,'' and ``frighteningly'' are hallucinated emotional states that have no signature in the joint coordinates. Yet standard contrastive objectives treat each draw as \emph{the} ground-truth target. No existing retrieval method models this many-to-one structure; the positive key $\mathbf{k}_+$---the text embedding that the contrastive loss pulls toward the motion query---is implicitly assumed to be deterministic when it is in fact stochastic.

The problem compounds across datasets. HumanML3D~\cite{guo2022humanml3d} and KIT-ML~\cite{plappert2016kit} describe overlapping motions but with different noise distributions: HumanML3D captions are verbose and intent-laden, while KIT-ML uses terse templates. A model trained on one noise distribution is miscalibrated for another---not because the motions differ, but because the \emph{corruption process} differs. Cross-dataset failure is a direct consequence of fitting to dataset-specific supervision noise rather than to the shared latent motion semantics.

We propose \textbf{MoCHA} (Motion Canonicalization for Human Action retrieval), a supervision denoising framework that projects captions onto their motion-recoverable content before they enter the contrastive objective. We formalize caption noise as a decomposition into motion-recoverable semantics $s$ and nuisance factors $a$, which capture annotator-dependent variation such as linguistic style and inferred context. We then define a canonicalization operator $C(t)$ that removes $a$ from the input caption $t$. The operator is first implemented via an LLM (GPT-5.2), then distilled into a lightweight FlanT5-base model that requires no LLM at inference time. Because MoCHA operates entirely on the text supervision channel, it is compatible with any retrieval architecture. We introduce \textbf{blend training}, a dual-pass contrastive scheme that balances denoised and original caption views during training, avoiding over-canonicalization while retaining the denoised alignment signal.

% v2
Our contributions are:
\begin{enumerate}[nosep,leftmargin=*]
    \item We define a text canonicalization operator $C(t)$ that projects captions onto motion-recoverable content $s$, stripping annotator-specific nuisance $a$. Canonicalization is a general principle: even rule-based variants improve retrieval. $C$ is implemented via GPT-5.2 and distilled into FlanT5 for LLM-free inference.

    \item We propose MoCHA with blend training, which balances canonicalized and original captions during contrastive learning. MoCHA sets a new state of the art on HumanML3D (H), KIT-ML (K), and cross-dataset H$\to$K retrieval (+3.1pp/+10.3pp T2M R@1 on H/K).

    \item We show that denoising supervision outperforms augmenting it: paraphrase augmentation widens $p(t|s)$ and can degrade R@1, while canonicalization consistently improves all ranks by collapsing within-motion embedding variance by 11--19\%.

    \item We provide empirical evidence that caption supervision contains systematic non-kinematic noise: canonicalization produces tighter positive clusters, better text-motion alignment, and improved separation from negatives in the learned embedding space.
\end{enumerate}

% % v1
% Our contributions are:
% \begin{enumerate}[nosep,leftmargin=*]
%     \item A text canonicalization operator that denoises caption supervision by projecting onto motion-recoverable content, with two variants---LLM-based (GPT-5.2) and a distilled FlanT5 for LLM-free inference. Crucially, canonicalization is a general principle: even deterministic rule-based methods (stopword stripping) improve cross-dataset transfer, establishing that the gains stem from noise removal itself, not LLM-specific language understanding.
%     \item MoCHA with blend training sets a new state of the art on HumanML3D, KIT-ML, and cross-dataset H$\to$K retrieval, achieving +3.1pp / +10.3pp T2M R@1 on H / K (LLM variant); +2.5pp / +8.1pp (T5, no LLM at inference).
%     \item Evidence that denoising outperforms augmentation: variance reduction yields +94\% cross-dataset transfer on H$\to$K, while paraphrase augmentation can \emph{hurt} in-distribution performance.
%     \item Empirical evidence that motion-caption supervision contains systematic non-kinematic noise: canonicalization reduces within-motion embedding variance by 11--19\%, confirmed as non-discriminative by uniform improvements across all recall ranks (R@1, R@5, R@10).
% \end{enumerate}

%==============================================================================
\section{Related Work}
\label{sec:related}
%==============================================================================

\paragraph{Motion--language learning.}
Text-conditioned motion generation has progressed from VAE-based models~\cite{petrovich2022temos,petrovich2021actor} through diffusion~\cite{tevet2023mdm,chen2023mld}, discrete-token~\cite{zhang2023t2mgpt,jiang2023motiongpt}, and masked-modeling~\cite{guo2024momask} approaches, with retrieval typically serving as an auxiliary evaluation.
TMR~\cite{petrovich2023tmr} formalized \emph{text-to-motion retrieval} as a standalone task via contrastive training, and MotionPatches (MoPa)~\cite{motionpatches} later showed that ViT-style patch tokenization with InfoNCE can outperform heavier generative formulations.
CLIP~\cite{radford2021clip} provides a widely reused text-embedding prior; MotionCLIP~\cite{tevet2022motionclip} and LaMP~\cite{li2025lamp} adapt vision--language pretraining to the motion domain.

\paragraph{Motion--language datasets.}
HumanML3D~\cite{guo2022humanml3d} (${\sim}$15k motions, ${\sim}$45k captions) and KIT-ML~\cite{plappert2016kit} (${\sim}$3.9k motions, ${\sim}$6.3k captions) are the primary benchmarks, sharing AMASS~\cite{mahmood2019amass} motion capture but differing sharply in annotation style: HumanML3D uses crowd-sourced free-form sentences while KIT-ML uses semi-templated descriptions.
BABEL~\cite{punnakkal2021babel} provides short action labels for ${\sim}$43k temporal segments, offering broad coverage but minimal linguistic detail.
These stylistic differences---verbosity, templating, granularity---create a domain gap that confounds cross-dataset evaluation even when the underlying motions overlap.

\paragraph{Cross-dataset transfer.}
TMR++~\cite{bensabath2024crossdataset} established cross-dataset protocols and showed that LLM paraphrases can partially reduce the annotation gap, but substantial bias remains.
MTR-MSE~\cite{xu2025mtrmse} expands motion semantics with LLM-generated descriptions, while LAVIMO~\cite{yin2024lavimo} adds video as a bridging modality.
These methods increase caption coverage without addressing the non-kinematic noise (hallucinated intent, annotator style) that drives the gap.

\paragraph{Text normalization.}
Diversifying supervision through paraphrases, back-translation, or LLM rewrites~\cite{bensabath2024crossdataset,xu2025mtrmse} increases surface-form coverage but can amplify dataset-specific style.
Text canonicalization has a long history in NLP~\cite{ji2022survey} and in vision--language retrieval via prompt templates.
Our work takes the opposite direction: rather than augmenting captions, we \emph{denoise} them by projecting onto motion-recoverable content (body parts, directions, timing), creating a shared canonical space that suppresses annotator-specific phrasing.

\section{Method}
\label{sec:method}
%==============================================================================

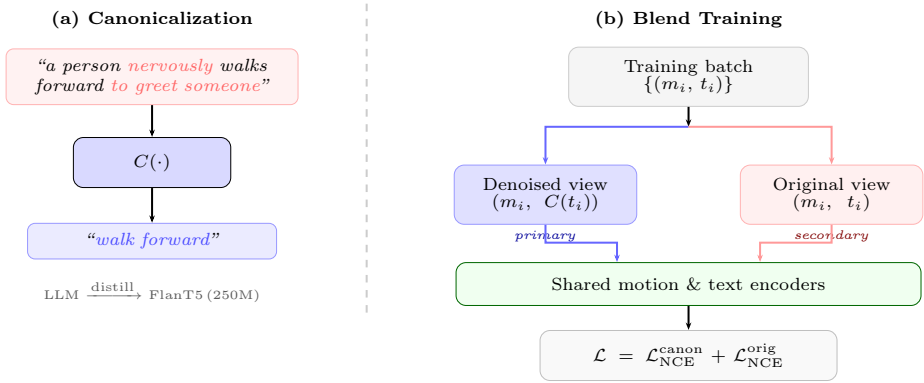
\begin{figure}[!t]
\centering
\footnotesize
\resizebox{\textwidth}{!}{%
\begin{tikzpicture}[
  block/.style={draw, rounded corners=4pt, minimum height=0.7cm, inner sep=5pt, font=\scriptsize},
  textbox/.style={font=\scriptsize, align=center},
  arr/.style={-{Stealth[length=3pt]}, thick},
  lbl/.style={font=\tiny, text=gray!70!black},
  dasharr/.style={-{Stealth[length=3pt]}, thick, dashed},
]

% === (a) Canonicalization ===
\node[font=\scriptsize\bfseries] at (0,2.6) {(a) Canonicalization};

% Input
\node[textbox, text width=3.8cm, fill=red!5, draw=red!30, rounded corners=3pt, inner sep=4pt] (input) at (0,1.8) {\textit{``a person \textcolor{red!60}{nervously} walks forward \textcolor{red!60}{to greet someone}''}};

% C(.) operator
\node[block, fill=blue!15, minimum width=2.2cm, font=\scriptsize\bfseries] (cop) at (0,0.6) {$C(\cdot)$};

% Output
\node[textbox, text width=3.2cm, fill=blue!8, draw=blue!40, rounded corners=3pt, inner sep=4pt] (output) at (0,-0.5) {\textit{``\textcolor{blue!70}{walk forward}''}};

% Arrows
\draw[arr] (input) -- (cop);
\draw[arr] (cop) -- (output);

% Implementation note
\node[lbl, text width=3.5cm, align=center] at (0,-1.2) {LLM $\xrightarrow{\text{distill}}$ FlanT5\,(250M)};

% Separator
\draw[gray!40, dashed, thick] (3,-1.5) -- (3,2.9);

% === (b) Blend Training ===
\node[font=\scriptsize\bfseries] at (7.5,2.6) {(b) Blend Training};

% Batch at top
\node[block, fill=gray!8, draw=gray!40, text width=3cm, align=center] (batch) at (7.5,1.8) {Training batch\\[-1pt]$\{(m_i,\, t_i)\}$};

% Fork arrows
\coordinate (fork) at (7.5,1.1);
\draw[arr] (batch) -- (fork);

% Denoised branch (left)
\node[block, fill=blue!12, draw=blue!40, text width=2.2cm, align=center] (denoised) at (5.5,0.15) {Denoised view\\[-1pt]$(m_i,\; C(t_i))$};
\node[lbl, below=-1pt of denoised, text=blue!60!black] {\textit{primary}};

% Original branch (right)
\node[block, fill=red!6, draw=red!30, text width=2.2cm, align=center] (original) at (9.5,0.15) {Original view\\[-1pt]$(m_i,\; t_i)$};
\node[lbl, below=-1pt of original, text=red!50!black] {\textit{secondary}};

% Fork lines
\draw[arr, blue!60] (fork) -| (denoised.north);
\draw[arr, red!40] (fork) -| (original.north);

% Shared encoder
\node[block, fill=green!6, draw=green!40!black, minimum width=6.5cm, minimum height=0.6cm, align=center] (encoder) at (7.5,-1.1) {Shared motion \& text encoders};
\draw[arr, blue!60] (denoised.south) -- ++(0,-0.25) -| ([xshift=-1cm]encoder.north);
\draw[arr, red!40] (original.south) -- ++(0,-0.25) -| ([xshift=1cm]encoder.north);

% Combined loss
\node[block, fill=gray!5, draw=gray!50, text width=3.8cm, align=center] (loss) at (7.5,-2.1) {$\mathcal{L} = \mathcal{L}_{\text{NCE}}^{\text{canon}} + \mathcal{L}_{\text{NCE}}^{\text{orig}}$};
\draw[arr] (encoder) -- (loss);

\end{tikzpicture}%
}
\caption{\textbf{MoCHA overview.} (a)~Motivated by the $(s, a)$ decomposition (Section~\ref{sec:theory}), $C(\cdot)$ projects each caption onto~$s$ by stripping stylistic variation~$a$ (\textcolor{red!60}{red}). $C$ is implemented via LLM and distilled into FlanT5 for LLM-free inference. (b)~Blend training balances both views: the denoised $C(t_i)$ anchors embeddings around~$s$ to reduce gradient variance, while the original~$t_i$ regularizes for natural-language queries.}
\label{fig:system}
\end{figure}

MoCHA is a supervision denoising framework for motion--text retrieval that operates entirely on the text channel, leaving the motion and text encoders unchanged. Given a motion sequence $m$ (e.g., a 3D joint-coordinate stream) and a natural-language caption $t$, standard retrieval models learn a motion encoder $M(\cdot)$ and a text encoder $T(\cdot)$ that map each modality into a shared embedding space, trained with an InfoNCE/CLIP-style contrastive objective over paired samples $(m_i, t_i)$.

%----------------------------------------------------------------------
\subsection{Contrastive Objectives Under Noisy Caption Supervision}
\label{sec:theory}
%----------------------------------------------------------------------

% Para 1
\paragraph{Captions mix motion semantics with nuisance factors.}
While a caption $t$ describes a motion $m$, it is not a deterministic label. Instead, it may be modeled as a noisy textual view that combines (i) \emph{motion-recoverable semantics} $s$ that depend on the 3D joint coordinates (e.g., action type, involved body parts, directionality, repetition counts), and (ii) \emph{nuisance} factors $a$ that capture annotator-dependent variation and are orthogonal to the motion itself. We use $a$ to subsume both linguistic variation (verbosity, syntax, paraphrase, filler words) and extra inferred context that a caption may add (assumed intent, objects, or purpose). Formally,
\begin{equation}
t \sim p(t \mid s, a), \qquad a \sim p(a).
\label{eq:caption_gen}
\end{equation}
For example, in ``a person walks forward to greet someone,'' the motion-recoverable semantics include ``walk forward" (part of $s$), while ``to greet someone'' is an inferred addition (part of $a$) whose truth is typically not determined by 3D joints alone. Crucially, often there is a large variety of plausible contexts for a given motion sequence, leading to significant noise in the resulting captions.

% v2 Para 2 (v1 Paras 2-4)
\paragraph{Distributional positives induce noisy supervision.}
If multiple annotators describe the same motion semantics $s$, the resulting captions form a distribution
\begin{equation}
p(t\mid s) = \int p(t\mid s,a)\,p(a)\,\mathrm{d}a .
\label{eq:prec}
\end{equation}
Thus each caption paired with a motion is effectively a random draw from $p(t\mid s)$, even though standard InfoNCE treats it as the single positive target, introducing stochastic variation in the supervision signal. In datasets with multiple captions per motion, the embeddings $\{T(t_k)\}$ therefore form a spread around a motion-dependent mean, producing nontrivial within-motion variance $\operatorname{Var}[T(t)\mid s]$. Contrastive training must align $M(m)$ to this varying target, weakening the specificity of the alignment signal. Empirically, we quantify this effect in Section~\ref{sec:variance_empirical}: our canonicalization reduces within-motion text-embedding variance by 11--19\%.

% Para 5
\paragraph{Distribution shift across datasets compounds the problem.}
Beyond within-dataset noise, the nuisance distribution $p(a)$ itself differs across datasets (e.g., verbosity, templating, propensity to add inferred context). Consequently, the expected positive text embedding $\mathbb{E}[T(t)\mid s]$ can shift even for identical motion semantics. A model trained under one dataset's annotation style is therefore systematically biased when evaluated against another dataset's caption distribution. Reducing sensitivity to $a$ is particularly important for cross-dataset transfer, where we observe the largest gains (H$\to$K +94\%, K$\to$H +52\%).

% Para 6
\paragraph{Why diversification alone may not help.}
Paraphrase-based augmentation typically increases linguistic variability while preserving the same inferred additions, effectively resampling nuisance factors rather than suppressing them. This widens $p(t\mid s)$ and can increase the spread of $T(t)\mid s$, which can hurt in-distribution retrieval (Section~\ref{sec:var_reduction}). MoCHA instead targets \emph{denoising}: it seeks a stable textual representation that preserves $s$ while suppressing $a$.

\subsection{Canonicalization as a Denoising Operator}
\label{sec:forward_canon}
%----------------------------------------------------------------------

% Para 1
Motivated by the analysis above, we seek a textual transformation that preserves motion-recoverable semantics $s$ while suppressing nuisance variation $a$. We therefore introduce a \emph{canonicalization operator} $C(\cdot)$ that maps a natural-language caption $t$ to a canonical form:
\begin{equation}
C(t) \approx \phi(s),
\end{equation}
where $\phi(s)$ denotes a textual representation that depends primarily on the underlying motion semantics and is invariant to annotator-specific variation. Intuitively, $C(\cdot)$ projects captions onto a stable description of the motion content while removing stylistic variation and extraneous inferred context.

% Para 2
Under this transformation, captions describing the same motion map to similar canonical forms, tightening the conditional embedding distribution so that $\operatorname{Var}[T(C(t)) \mid s] < \operatorname{Var}[T(t) \mid s]$. Canonical captions therefore provide a more stable supervision signal for contrastive alignment.

% Para 3
\paragraph{LLM-based canonicalization.}
We implement $C(\cdot)$ using a large language model prompted with a small number of examples that illustrate how to extract motion-recoverable semantics while discarding nuisance elements. The prompt instructs the model to preserve only motion-relevant content such as actions, body parts, directions, and repetitions, while removing stylistic phrasing and inferred context. The full prompt specification is provided in Appendix~\ref{app:llm_prompt}.

% Para 4
\paragraph{Distilled canonicalizer.}
To eliminate dependence on an external LLM at deployment and reduce inference latency, we distill the canonicalization operator into a compact model. Specifically, we train a FlanT5-base model~\cite{chung2022flant5} on pairs $\{(t, C(t))\}$ generated by the LLM, enabling efficient canonicalization at both training and inference time. Implementation details are provided in Appendix~\ref{app:implementation}.

%----------------------------------------------------------------------
\subsection{Blend Training}
\label{sec:blend_training}
%----------------------------------------------------------------------

To balance semantic stability with linguistic diversity, we introduce \emph{Blend Training}, which combines canonicalized and original captions during optimization. For each batch $\{(m_i, t_i)\}$, we compute two contrastive objectives:
\begin{equation}
\mathcal{L}_{\mathrm{mix}}
=
\lambda\,\mathcal{L}_{\mathrm{InfoNCE}}\!\left(\{(m_i, C(t_i))\}\right)
+
(1-\lambda)\,\mathcal{L}_{\mathrm{InfoNCE}}\!\left(\{(m_i, t_i)\}\right),
\label{eq:Lmix}
\end{equation}
where the canonical term provides low-variance supervision centered on motion semantics $s$, and the original-caption term exposes the model to the broader caption distribution.

This combination is useful for two reasons. First, canonicalization can occasionally compress or omit motion-relevant cues present in the original caption; retaining the original view preserves such signals. Second, while canonical captions stabilize the supervision signal, original captions maintain linguistic diversity that acts as a regularizer, preventing the encoder from overfitting to a single canonical phrasing. Together, the two views provide complementary supervision: canonical captions anchor alignment around $s$, while original captions sample the broader caption distribution.

Operationally, we implement this as two sequential passes per batch through shared motion and text encoders. In our default configuration, \textbf{Blend-Rev}, the canonical pass is applied first to establish semantic alignment, followed by the original-caption pass as regularization. This simple strategy allows MoCHA to benefit from denoised supervision while remaining robust to natural-language queries at inference time.

\section{Empirical Analysis}
\label{sec:analysis}
%==============================================================================

We first validate the noisy-supervision formulation from Section~\ref{sec:theory} with two complementary analyses: direct measurement of supervision noise reduction (Section~\ref{sec:variance_empirical}) and its effect on embedding space geometry (Section~\ref{sec:embedding_geometry}). We then present retrieval results in Section~\ref{sec:experiments}. All experiments use HumanML3D (H3D)~\cite{guo2022humanml3d} and KIT-ML (KIT)~\cite{plappert2016kit}. Additional BABEL~\cite{punnakkal2021babel} results are in Appendix~\ref{app:crossds_babel}.

%----------------------------------------------------------------------
\subsection{Setup}
\label{sec:setup}
%----------------------------------------------------------------------

\paragraph{Architecture.} We use MotionPatches (MoPa)~\cite{motionpatches} as the primary retrieval model: ViT-B/16 over 22-joint 3D representations, DistilBERT CLS-pooled text encoder, 256-dim shared space, frozen temperature $\tau = 0.07$, cosine LR schedule with per-batch stepping, no gradient clipping. For HumanML3D, the motion encoder uses a 10$\times$ lower learning rate. All models train for 50 epochs with batch size 128 and base LR $10^{-5}$.

\paragraph{Canonicalization variants.} We report two MoCHA variants: \textbf{MoCHA (LLM)}, which uses GPT-5.2 canonicalization at both train and test, and \textbf{MoCHA (T5)}, a fully LLM-free variant using FlanT5-PPT at both train and test. The FlanT5 canonicalizer is trained on 168K pairs: $\sim$130K TMR++ paraphrases~\cite{bensabath2024crossdataset} filtered to HumanML3D and KIT-ML training splits (paired with their LLM canonicalizations), plus $\sim$38K original training captions paired with their LLM canonicals. Full variant details and ablations are in Appendix~\ref{app:strategy_comparison}.

\paragraph{Evaluation.} All results use the DsPair protocol~\cite{motionpatches}, which evaluates retrieval over the full test set and credits a match to any caption belonging to the same motion---providing a stricter measure than threshold-based grouping (Appendix~\ref{app:protocol_definitions}). We report T2M and M2T R@1, R@5, R@10.

%----------------------------------------------------------------------
\subsection{Measuring Supervision Noise}
\label{sec:variance_empirical}
%----------------------------------------------------------------------

Section~\ref{sec:theory} argued that caption supervision injects noise into the positive key $\mathbf{k}_+$ through annotator-specific draws of the nuisance factors $a$. Multi-caption datasets allow us to measure this effect directly. Each motion in HumanML3D and KIT-ML has $K \geq 3$ captions written independently by annotators watching the same motion clip. These captions are not paraphrases---they reflect genuine annotator variation in what aspects of the motion to describe (e.g., ``a person walks forward nervously,'' ``someone takes careful steps ahead,'' ``walk forward slowly''). For each motion $m$, we embed all $K$ captions with the baseline text encoder and compute the mean pairwise cosine dissimilarity $V(m) = 1 - \frac{1}{\binom{K}{2}} \sum_{i < j} \cos(\mathbf{t}_i, \mathbf{t}_j)$ as a measure of within-motion embedding spread. We then canonicalize all $K$ captions---removing annotator-specific style and inferred context to isolate the shared motion content (e.g., all three $\rightarrow$ ``walk forward'')---embed the canonical versions through the same encoder, and recompute the variance.

\begin{table}[!ht]
\centering
\caption{\textbf{Within-motion text embedding variance} (C4). Canonicalization reduces the spread of caption embeddings for the same motion by 11--19\%, confirming that $a$ introduces measurable noise into the contrastive positive. $V(m)$: mean pairwise cosine dissimilarity across a motion's $K$ captions in the baseline encoder's 256-dim space.}
\label{tab:variance_reduction}
\scriptsize
\setlength{\tabcolsep}{4pt}
\begin{tabular}{l c ccc}
\toprule
\textbf{Dataset} & $N$ & Original & MoCHA (T5) & $\Delta$ \\
\midrule
HumanML3D & 4{,}344 & 0.587 & 0.522 & $-$11.1\% \\
KIT-ML    & 210    & 0.561 & 0.456 & $-$18.7\% \\
\bottomrule
\end{tabular}
\end{table}

$V(m)$ directly measures how much the positive key $\mathbf{k}_+$ varies due to caption selection. During training, each step samples one of the $K$ captions for motion $m$ and encodes it as $\mathbf{k}_+ = T(c_k)$. For $L^2$-normalized embeddings, the text-selection variance decomposes as
\begin{equation}
\mathrm{Var}_{\mathrm{text}}[\mathbf{k}_+]
\;=\; \frac{1}{K}\sum_{k=1}^{K} \lVert T(c_k) - \bar{\mathbf{t}}_m \rVert^2
\;=\; \frac{K{-}1}{K}\,V(m),
\label{eq:var_text}
\end{equation}
where $\bar{\mathbf{t}}_m = \frac{1}{K}\sum_k T(c_k)$ is the embedding centroid. Table~\ref{tab:variance_reduction} confirms the prediction from Section~\ref{sec:theory}: canonicalization reduces $V(m)$ by 11\% on HumanML3D and 19\% on KIT-ML (both $p < 10^{-6}$). The larger reduction on KIT-ML likely reflects its greater caption diversity (mean 4 captions/motion vs.\ H3D's mean 3). In effect, independent annotators' descriptions of the same motion \emph{converge} once non-motion content is removed.

\paragraph{Direct gradient variance measurement.}
Reducing $V(m)$ implies that the gradient signal for a motion should also become more consistent across caption choices. To test this, we measure InfoNCE gradient variance on a frozen baseline model (trained only on original captions, epoch 41). For each motion, we compute gradients once using original captions and once using FlanT5 canonical captions across all 4{,}382 multi-caption test motions, isolating the effect of caption choice under a fixed model.

Canonical captions reduce gradient variance $\sigma^2$ by 11.1\% (79.72 $\rightarrow$ 70.89), with 69.5\% of individual motions showing lower variance. The gradient cone width—the angular spread of per-caption gradients for the same motion—shrinks by 7.4\% (38.1$^\circ$ $\rightarrow$ 35.3$^\circ$), while gradient cosine consistency increases by 30.2\%. Notably, the 11\% reduction in $\sigma^2$ closely matches the 11\% reduction in $V(m)$ from Table~\ref{tab:variance_reduction}, indicating that input-level variance reduction propagates directly to gradient-level noise reduction.

Lower gradient variance leads to a sharper contrastive signal during training. Consistent with this expectation, MoCHA-trained models produce more concentrated InfoNCE distributions, with lower softmax entropy (6.03 vs.\ 6.29 baseline on H3D) and 21\% higher probability assigned to the correct positive (Appendix~\ref{app:gradient_concentration}). Together, these results support the first part of the causal chain proposed in Section~\ref{sec:theory}: canonicalization reduces caption-induced supervision noise, which leads to more consistent gradients during contrastive training. We next examine how this cleaner supervision affects the geometry of the learned embedding space.

\subsection{Embedding Space Geometry}
\label{sec:embedding_geometry}
%----------------------------------------------------------------------

We analyze the geometry of the learned 256-dimensional embedding space. For each multi-caption motion in the test set, we compute four metrics: \textbf{Intra Sim}---average cosine similarity between text embeddings of captions describing the same motion; \textbf{Text-Motion Align}---average cosine similarity between each text embedding and its paired motion embedding; \textbf{Inter NN Sim}---average cosine similarity to the nearest \emph{negative neighbor} (a caption describing a different motion); and \textbf{Sep Ratio}---Intra Sim / Inter NN Sim, measuring the separation between same-motion captions and their nearest negatives (higher is better).

\begin{table}[!ht]
\centering
\caption{\textbf{Embedding space geometry} (C4): baseline vs.\ MoCHA in the trained encoder's 256-dim space. By removing $a$ from the training signal, MoCHA produces embeddings where same-motion captions cluster more tightly (Intra) and align more closely with their motion (Align), improving separation by +8--25\%. This confirms that input-level variance reduction propagates to a better-structured retrieval space. Sep: Intra / Inter NN. Full breakdown in Appendix~\ref{app:embedding_geometry_full}.}
\label{tab:embedding_geometry}
\scriptsize
\setlength{\tabcolsep}{2.5pt}
\begin{tabular}{ll cccc}
\toprule
& \textbf{Model} & Intra & Align & Inter NN & Sep \\
\midrule
\multicolumn{6}{l}{\textit{HumanML3D}} \\
\quad & Baseline   & 0.413 & 0.601 & 0.976 & 0.423 \\
\quad & MoCHA      & \best{0.444} & \best{0.636} & \best{0.971} & \best{0.457} \\
\midrule
\multicolumn{6}{l}{\textit{KIT-ML}} \\
\quad & Baseline   & 0.456 & 0.515 & 0.941 & 0.484 \\
\quad & MoCHA      & \best{0.566} & \best{0.553} & \best{0.938} & \best{0.604} \\
\bottomrule
\end{tabular}
\end{table}

Table~\ref{tab:embedding_geometry} compares the embedding space geometry of the baseline (trained on original captions) and MoCHA (Blend-Rev trained, canonical evaluation). On both datasets, MoCHA produces tighter positive clusters (higher Intra Sim), stronger text–motion alignment, and improved separation from negatives. On H3D, intra similarity increases from 0.413 to 0.444 (+7.5\%) and alignment from 0.601 to 0.636 (+5.8\%), while the separation ratio improves from 0.423 to 0.457 (+8\%). The gains are even larger on KIT-ML: intra similarity +24\%, alignment +7.4\%, and separation +25\%.

Importantly, tighter positive clusters do \emph{not} reduce inter-class separation. The similarity to the nearest negative caption slightly decreases (e.g., 0.976 $\rightarrow$ 0.971 on H3D), indicating that negatives are pushed further away rather than collapsed toward the positives. Overall, these results show that canonicalization produces a more structured embedding space: captions describing the same motion cluster more tightly, align more closely with their paired motions, and remain well separated from captions describing different motions.

\section{Retrieval Results}
\label{sec:experiments}
%==============================================================================

\begin{figure*}[!t]
\centering
%--- Row 1 ---
\begin{minipage}[t]{0.48\textwidth}
\centering
\textbf{(a)} Query: \textit{``a figure leans forward as though impersonating an airplane''}\\
\textbf{MoCHA:} \textit{``lean forward (as if imitate airplane)''}\quad Rank: 420\,$\to$\,\textbf{1}\\[2pt]
\includegraphics[width=\textwidth,trim=0 15 0 10,clip]{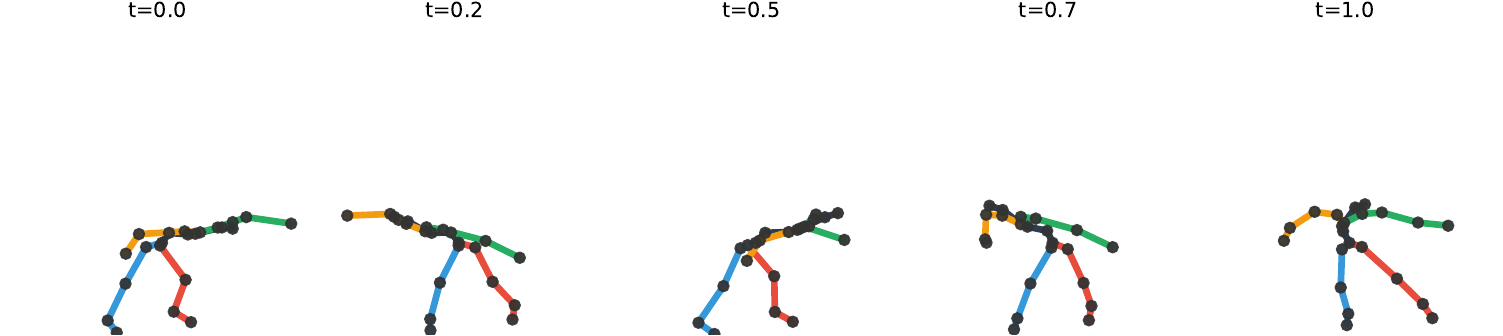}\\[-4pt]
{\scriptsize\color{gray} Correct motion (MoCHA rank 1)}\\[1pt]
\includegraphics[width=\textwidth,trim=0 15 0 10,clip]{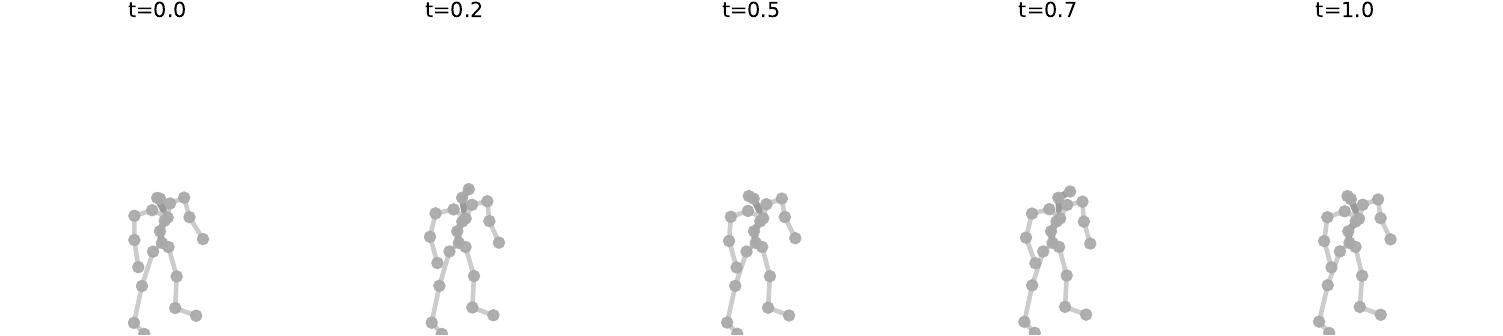}\\[-4pt]
{\scriptsize\color{gray} Baseline rank-1 retrieval (standing head rotation)}
\end{minipage}%
\hfill
\begin{minipage}[t]{0.48\textwidth}
\centering
\textbf{(b)} Query: \textit{``figure appears to be fighting or dancing''}\\
\textbf{MoCHA:} \textit{``dance''}\quad Rank: 211\,$\to$\,\textbf{1}\\[2pt]
\includegraphics[width=\textwidth,trim=0 15 0 10,clip]{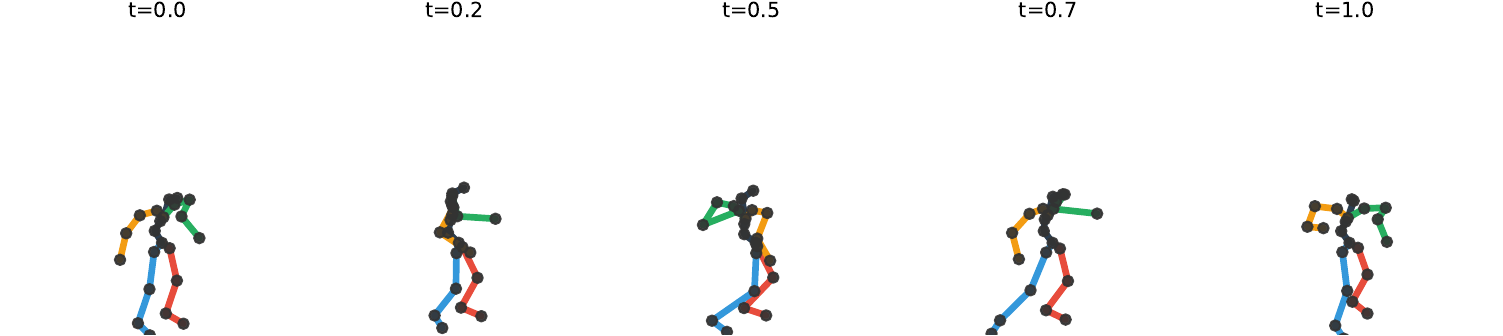}\\[-4pt]
{\scriptsize\color{gray} Correct motion (MoCHA rank 1)}\\[1pt]
\includegraphics[width=\textwidth,trim=0 15 0 10,clip]{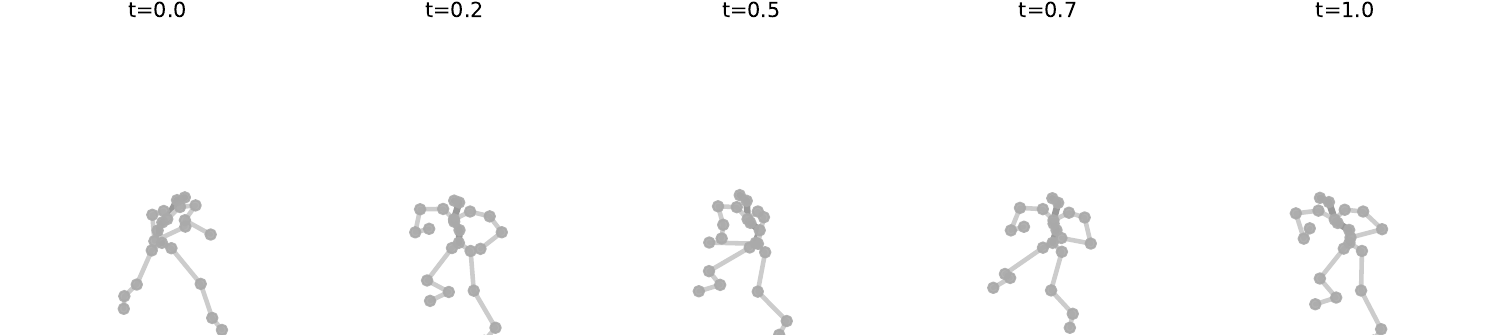}\\[-4pt]
{\scriptsize\color{gray} Baseline rank-1 retrieval (boxing jabs)}
\end{minipage}

\vspace{3pt}
%--- Row 2 ---
\begin{minipage}[t]{0.48\textwidth}
\centering
\textbf{(c)} Query: \textit{``a person walks forward with exaggerated backward kicks''}\\
\textbf{MoCHA:} \textit{``walk forward $\to$ kick back exaggeratedly each step''}\quad Rank: 402\,$\to$\,\textbf{2}\\[2pt]
\includegraphics[width=\textwidth,trim=0 15 0 10,clip]{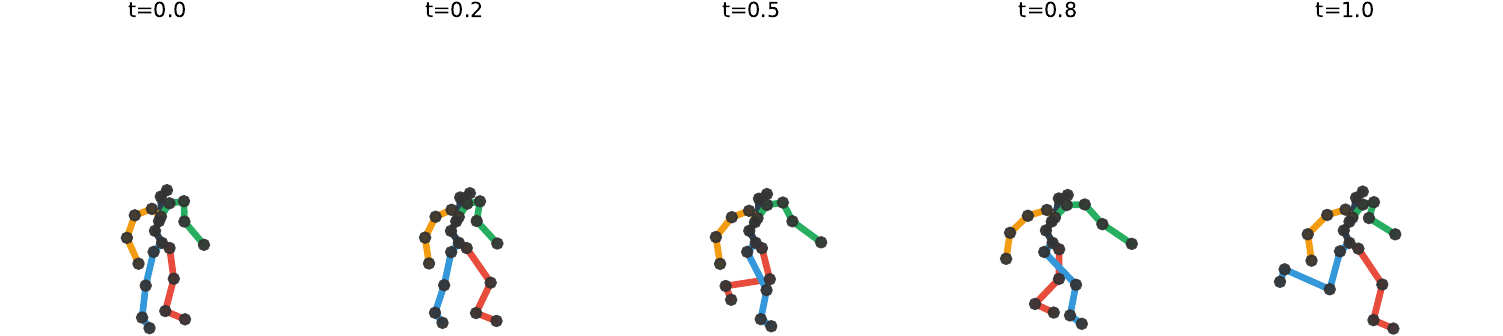}\\[-4pt]
{\scriptsize\color{gray} Correct motion (MoCHA rank 2)}\\[1pt]
\includegraphics[width=\textwidth,trim=0 15 0 10,clip]{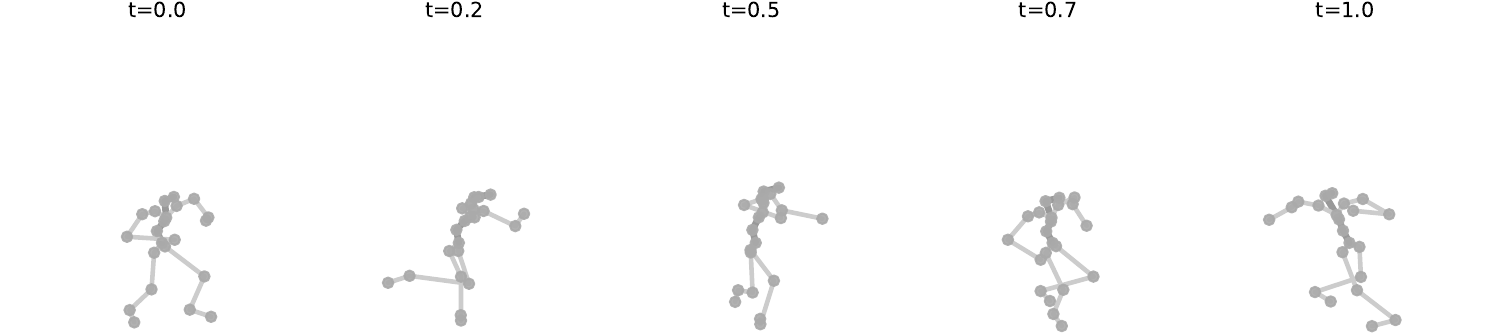}\\[-4pt]
{\scriptsize\color{gray} Baseline rank-1 retrieval (left/right leg kicks)}
\end{minipage}%
\hfill
\begin{minipage}[t]{0.48\textwidth}
\centering
\textbf{(d)} Query: \textit{``a person walking in a straight path at a slow pace with only four steps''}\\
\textbf{MoCHA:} \textit{``walk straight slow $\to$ walk four steps arms at sides''}\quad Rank: 557\,$\to$\,\textbf{1}\\[2pt]
\includegraphics[width=\textwidth,trim=0 15 0 10,clip]{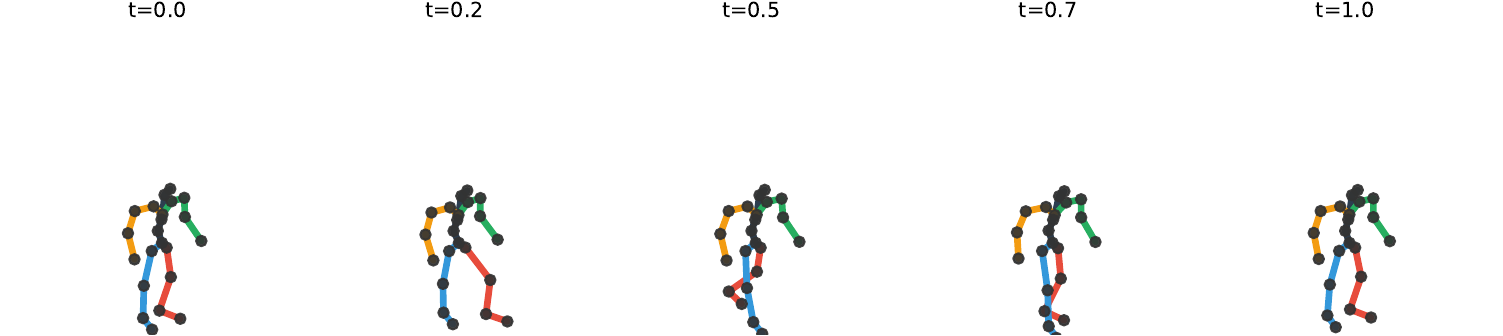}\\[-4pt]
{\scriptsize\color{gray} Correct motion (MoCHA rank 1)}\\[1pt]
\includegraphics[width=\textwidth,trim=0 15 0 10,clip]{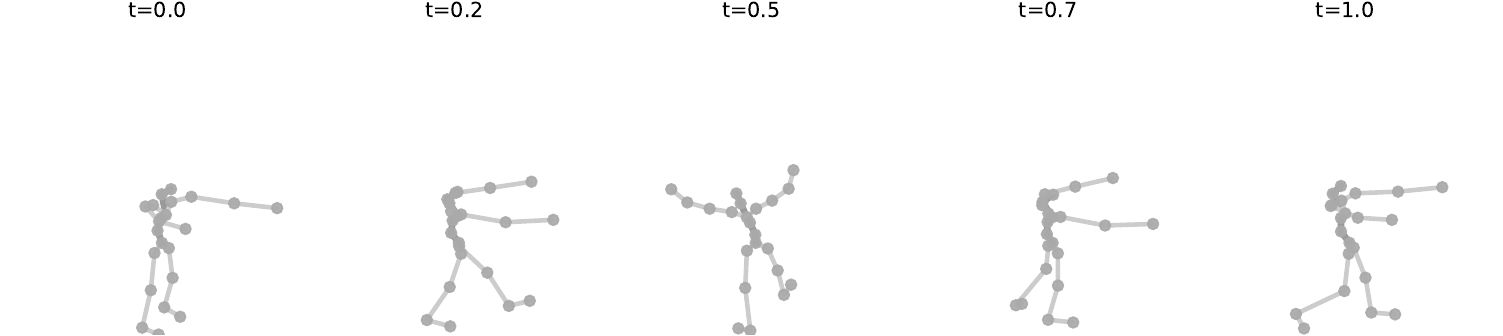}\\[-4pt]
{\scriptsize\color{gray} Baseline rank-1 retrieval (walking arms forward)}
\end{minipage}
\caption{\textbf{Canonicalization projects captions onto $s$, improving retrieval.} Top row (colored): ground truth; bottom row (gray): baseline rank-1 error. (a)~Verbose $a$ buries the action; MoCHA extracts $s$ while preserving the metaphor. (b)~Annotator uncertainty ($a$); canonicalization extracts shared kinematic content. (c)~Complex description decomposed into sequential $s$, disambiguating from similar motions. (d)~Over-specified caption dilutes the contrastive signal; MoCHA strips $a$, retains discriminative $s$.}
\label{fig:retrieval_improvements}
\end{figure*}

Having established the theoretical basis for caption denoising (Section~\ref{sec:theory}) and empirically validated its effect on supervision noise and embedding geometry (Section~\ref{sec:analysis}), we now evaluate the end-to-end retrieval performance of MoCHA. We begin with in-distribution benchmarks, where MoCHA achieves state-of-the-art on both HumanML3D and KIT-ML (\S\ref{sec:indist}). We then show that canonicalization is especially effective for cross-dataset transfer, where removing dataset-specific annotation style yields up to +94\% relative improvement (\S\ref{sec:crossds}). Finally, we demonstrate that denoising outperforms paraphrase augmentation (\S\ref{sec:var_reduction}) and that canonicalization is a general principle---even simple rule-based methods improve transfer, though learned canonicalizers like MoCHA provide substantially larger gains (\S\ref{sec:rule_based_ablation}).

%----------------------------------------------------------------------
\subsection{In-Distribution Results}
\label{sec:indist}
%----------------------------------------------------------------------

\begin{table}[t]
\centering
\caption{\textbf{In-distribution retrieval results} (C2). MoCHA achieves state-of-the-art on both benchmarks, with consistent gains across all recall ranks and retrieval directions---ruling out a precision-recall tradeoff and confirming that the removed content was $a$, not useful $s$. Full ablations in Appendix~\ref{app:strategy_comparison}.}
\label{tab:indist_h3d}
\label{tab:indist_kit}
\scriptsize
\setlength{\tabcolsep}{2.5pt}
\begin{tabular}{l cccc cccc}
\toprule
& \multicolumn{4}{c}{\textbf{Text to motion}} & \multicolumn{4}{c}{\textbf{Motion to text}} \\
\cmidrule(lr){2-5} \cmidrule(lr){6-9}
\textbf{Methods} & R@1$\uparrow$ & R@5$\uparrow$ & R@10$\uparrow$ & MedR$\downarrow$ & R@1$\uparrow$ & R@5$\uparrow$ & R@10$\uparrow$ & MedR$\downarrow$ \\
\midrule
\multicolumn{9}{l}{\textit{HumanML3D}} \\
TEMOS & 2.12 & 8.26 & 13.52 & 173.0 & 3.86 & 9.38 & 14.00 & 183.25 \\
T2M & 1.80 & 7.12 & 12.47 & 81.00 & 2.92 & 8.36 & 12.95 & 81.50 \\
TMR & 8.92 & 22.06 & 33.37 & 25.00 & 9.44 & 22.92 & 32.21 & 26.00 \\
MoPa & 10.80 & 26.72 & 38.02 & 19.00 & 11.25 & 26.86 & 37.40 & 20.50 \\
\textbf{MoCHA (T5)} & \second{13.30} & \second{31.00} & \second{44.64} & \second{14.0} & \second{12.77} & \second{28.26} & \second{38.78} & \second{20.0} \\
\textbf{MoCHA (LLM)} & \best{13.91} & \best{33.53} & \best{45.14} & \best{13.0} & \best{14.37} & \best{30.43} & \best{40.67} & \best{17.0} \\
\midrule
\multicolumn{9}{l}{\textit{KIT-ML}} \\
TEMOS & 7.11 & 24.10 & 35.66 & 24.00 & 11.69 & 26.63 & 36.39 & 26.50 \\
T2M & 3.37 & 16.87 & 27.71 & 28.00 & 4.94 & 16.14 & 25.30 & 28.50 \\
TMR & 10.05 & 30.03 & 44.66 & 14.00 & 11.83 & 29.39 & 38.55 & 16.00 \\
MoPa & 14.02 & 34.10 & 50.00 & \second{10.50} & 13.61 & 33.33 & 44.77 & 13.00 \\
\textbf{MoCHA (T5)} & \second{22.14} & \second{47.84} & \second{59.67} & \best{6.0} & \second{18.32} & \best{36.90} & \second{46.69} & \second{12.0} \\
\textbf{MoCHA (LLM)} & \best{24.30} & \best{48.47} & \best{62.98} & \best{6.0} & \best{18.45} & \second{35.24} & \best{47.46} & \best{11.0} \\
\bottomrule
\end{tabular}
\end{table}

MoCHA substantially improves over MoPa on both benchmarks (Table~\ref{tab:indist_h3d}), achieving new state-of-the-art performance on HumanML3D and KIT-ML. Gains are larger on KIT-ML (+73\% relative improvement in R@1 for MoCHA-LLM, +58\% for MoCHA-T5) than on HumanML3D (+29\% and +23\%, respectively). The larger KIT-ML gains likely reflect its smaller dataset size and more templated annotation style, where caption noise constitutes a larger fraction of the supervision signal. Improvements are consistent across R@1, R@5, and R@10 for both retrieval directions, ruling out a precision–recall tradeoff and indicating that the removed caption content was introducing noise into the contrastive supervision rather than providing useful discriminative information. Full ablations appear in Appendix~\ref{app:strategy_comparison}.

%----------------------------------------------------------------------
\subsection{Cross-Dataset Results}
\label{sec:crossds}
%----------------------------------------------------------------------

\begin{table}[t]
\centering
\caption{\textbf{Cross-dataset retrieval.} Stripping $p(a)$ removes dataset-specific annotation style, so models trained on one dataset's conventions can retrieve from another's. Gains are proportionally larger than in-distribution---consistent with the $(s, a)$ decomposition in Section~\ref{sec:theory}, which predicts that $a$ hurts most when the test distribution of $a$ differs from training.}
\label{tab:crossds}
\scriptsize
\setlength{\tabcolsep}{2.5pt}
\begin{tabular}{ll l ccc}
\toprule
& & & \multicolumn{3}{c}{\textbf{T2M}} \\
\cmidrule(lr){4-6}
\textbf{Train} & \textbf{Model} & \textbf{Test} & R@1 & R@5 & R@10 \\
\midrule
\multirow{3}{*}{H3D}
& MoPa                  & KIT   & 13.74 & 37.79 & 53.31 \\
& MoCHA (T5 blend-rev)  & KIT   & \second{25.70} & \second{44.27} & \best{61.83} \\
& MoCHA (LLM blend-rev) & KIT   & \best{26.59} & \best{48.35} & \second{61.07} \\
\midrule
\multirow{3}{*}{KIT}
& MoPa                  & H3D   & 1.85 & 6.02 & 9.56 \\
& MoCHA (T5 blend)      & H3D   & \second{2.01} & \second{8.65} & \second{13.00} \\
& MoCHA (LLM blend)     & H3D   & \best{2.81} & \best{8.71} & \best{13.23} \\
\bottomrule
\end{tabular}
\end{table}

Cross-dataset retrieval provides a particularly strong test of the caption-denoising hypothesis because it removes any advantage from learning dataset-specific annotation style. Canonicalization produces large cross-dataset gains in both directions (Table~\ref{tab:crossds}), with H$\to$K improving by up to +94\% relative in R@1. This behavior is predicted by the analysis in Section~\ref{sec:theory}: the baseline learns dataset-specific annotation style $p(a)$, so it is miscalibrated when tested on a different dataset's conventions. Canonicalization strips this dataset-specific component, leaving representations that transfer. Embedding-space analysis further supports this interpretation: cross-dataset caption similarity increases by +12--14\% for matched motions after canonicalization (Appendix~\ref{app:crossds_caption_alignment}).

%----------------------------------------------------------------------
\subsection{Denoising the Positive vs.\ Augmenting It}
\label{sec:var_reduction}
%----------------------------------------------------------------------

TMR++~\cite{bensabath2024crossdataset} takes the opposite approach: rather than denoising captions, it augments them by substituting LLM-generated paraphrases during training. Section~\ref{sec:theory} predicts this should increase text-selection variance rather than reduce it.

\begin{table}[!ht]
\centering
\caption{\textbf{Denoising the positive vs.\ augmenting it.} Paraphrase augmentation widens $p(t\mid s)$ rather than reducing it, acting as a smoothing function that trades R@1 for R@5/R@10. Canonicalization improves all ranks by collapsing variance rather than adding to it.}
\label{tab:aug_vs_canon}
\scriptsize
\setlength{\tabcolsep}{2.5pt}
\begin{tabular}{ll ccc}
\toprule
& & \multicolumn{3}{c}{\textbf{T2M}} \\
\cmidrule(lr){3-5}
& & R@1 & R@5 & R@10 \\
\midrule
\multicolumn{5}{l}{\textit{H3D-trained $\to$ HumanML3D test}} \\
\quad & MoPa             & 10.80 & 26.72 & 38.02 \\
\quad & $+$Paraphrases   & 10.13 \loss{-0.67} & 27.62 & 40.31 \\
\quad & MoCHA (T5)       & \best{13.96} & \best{31.41} & \second{43.82} \\
\quad & MoCHA (LLM)      & \second{12.55} & \second{29.88} & \best{44.64} \\
\midrule
\multicolumn{5}{l}{\textit{H3D-trained $\to$ KIT-ML test (cross-dataset)}} \\
\quad & MoPa             & 13.74 & 37.79 & 53.31 \\
\quad & $+$Paraphrases   & 15.27 & 39.44 & 53.94 \\
\quad & MoCHA (T5)       & \second{25.70} & \second{44.27} & \best{61.83} \\
\quad & MoCHA (LLM)      & \best{26.59} & \best{48.35} & \second{61.07} \\
\midrule
\multicolumn{5}{l}{\textit{KIT-trained $\to$ KIT-ML test}} \\
\quad & MoPa             & 14.02 & 34.10 & 50.00 \\
\quad & $+$Paraphrases   & 11.45 \loss{-2.57} & 38.04 & 53.56 \\
\quad & MoCHA (T5)       & \second{17.05} & \second{41.09} & \best{62.21} \\
\quad & MoCHA (LLM)      & \best{24.05} & \best{48.22} & \best{62.21} \\
\bottomrule
\end{tabular}
\end{table}

Table~\ref{tab:aug_vs_canon} confirms this---paraphrase augmentation hurts in-distribution R@1 in two of three settings, because the model must match each motion to a moving target of diverse phrasings. MoCHA consistently improves, and the gap widens on cross-dataset transfer: canonicalization maps both datasets onto shared content $s$, whereas paraphrases only diversify captions within the source dataset's annotation style.

%----------------------------------------------------------------------
\subsection{Is Canonicalization a General Principle?}
\label{sec:rule_based_ablation}
%----------------------------------------------------------------------

To isolate what drives MoCHA's gains, we train two additional baselines under the same blend-rev protocol: \emph{stopword stripping} (remove determiners, pronouns, and discourse markers---the simplest form of denoising), and \emph{backtranslation} (EN$\to$DE$\to$EN via MarianMT, which paraphrases surface form while preserving stylistic properties---a negative control that transforms text without denoising it). Each method is trained and tested on its own transformed captions, eliminating train-test mismatch confounds.

\begin{table}[!ht]
\centering
\caption{\textbf{Canonicalization mechanism ablation} (T2M R@1~\%). Even rule-based stopword stripping improves transfer, while backtranslation (which transforms without denoising) does not---confirming that canonicalization is a general principle (C1): the gains stem from projecting onto $s$, not from any particular model's language understanding.}
\label{tab:rule_based}
\scriptsize
\setlength{\tabcolsep}{2.5pt}
\begin{tabular}{l cccc}
\toprule
& Baseline & Backtrans & Stopword & FT5 (MoCHA) \\
\midrule
H$\to$H & 10.80 & 8.71 & 10.58 & \best{13.12} \\
K$\to$K & 14.02 & 11.45 & 15.39 & \best{20.74} \\
H$\to$K & 13.74 & 15.39 & 16.67 & \best{18.96} \\
K$\to$H & 1.78 & 1.71 & 2.17 & \best{2.78} \\
\bottomrule
\end{tabular}
\end{table}

\begin{table}[!ht]
\centering
\caption{\textbf{Example outputs by method.} Backtranslation (negative control: transforms surface form without removing $a$) preserves nuisance factors intact (``as if you were following it''); stopword stripping removes noise indiscriminately, unlocking cross-dataset gains but losing temporal structure. MoCHA retains sequential actions and fine-grained $s$ that rule-based methods cannot distinguish from $a$---illustrating why canonicalization is a general principle that even simple rules can exploit, but learned semantic projection yields the largest improvements.}
\label{tab:canon_examples}
\scriptsize
\setlength{\tabcolsep}{2pt}
\begin{tabular}{@{}p{2.8cm} p{2.8cm} p{2.6cm} p{2.8cm}@{}}
\toprule
\textbf{Original} & \textbf{Backtrans} & \textbf{Stopword} & \textbf{FlanT5} \\
\midrule
``a person holding an item turns back, then left, then back, \textbf{like he's been followed}'' &
``a person holding an object turns back, then left, then back, \textbf{as if you were following it}'' &
``holding item turns back left back he's followed'' &
``hold item $\to$ turn back $\to$ turn left $\to$ turn back'' \\[6pt]
``a man holds something in both hands and walks in a counterclockwise oval, before coming to a stop, then resuming walking.'' &
``a man holds something in both hands and goes oval in counterclockwise sense before he comes to \textbf{an attack}, then goes again.'' &
``holds hands walks counterclockwise oval coming stop resuming walking'' &
``hold object both hands $\to$ walk counterclockwise oval $\to$ stop $\to$ walk'' \\
\bottomrule
\end{tabular}
\end{table}

Backtranslation confirms the mechanism: despite changing surface wording, it preserves annotation style---hallucinated intent (``like he's been followed'' $\to$ ``as if you were following it''), verbosity, and hedging---and produces no consistent benefit, performing at or below baseline in three of four conditions. This rules out the hypothesis that any text transformation helps; the active ingredient is specifically noise removal. Stopword stripping confirms the principle extends beyond learned canonicalization, improving over baseline on K$\to$K (+1.4pp), H$\to$K (+2.9pp), and K$\to$H (+0.4pp). However, it fails on H$\to$H where verbose captions interleave noise with discriminative detail that crude stripping cannot distinguish. FlanT5 is the only method that consistently improves all four conditions, because it strips annotator noise while preserving temporal structure and fine-grained motion content---a precision-recall tradeoff that simple rule-based methods cannot navigate.

\subsection{Qualitative Results}
\label{sec:qualitative}
%----------------------------------------------------------------------

Figure~\ref{fig:retrieval_improvements} shows T2M retrieval examples where canonicalization produces large rank improvements. Each example illustrates a different failure mode of noisy caption supervision: verbose phrasing~(a), ambiguous annotator intent~(b), implicit temporal structure~(c), and over-specification~(d). In all cases, the baseline's contrastive objective aligns to the full caption including nuisance factors~$a$, retrieving motions that match the noise rather than the motion-recoverable content~$s$. MoCHA projects each caption onto~$s$, producing a cleaner alignment target.

%==============================================================================
\section{Discussion}
\label{sec:discussion}
%==============================================================================

\paragraph{Canonicalization as a general principle.}
The rule-based ablation (Table~\ref{tab:rule_based}) establishes that gains stem from noise removal itself, not LLM-specific language understanding: even stopword stripping improves cross-dataset transfer, while backtranslation---which transforms text without removing annotation noise---provides no benefit, serving as a negative control. The gains are also architecture-agnostic, holding across both TMR and MoPa encoders (Appendix~\ref{app:tmr}).

\paragraph{Why blend training succeeds.}
Pure canonical training risks overcanonicalization: the LLM can strip detail that is genuinely part of $s$, collapsing semantically distinct motions onto the same target. The original-text pass acts as a regularizer, preserving fine-grained distinctions that the canonicalizer may discard. This dual-pass scheme denoises the primary alignment signal while guarding against information loss (Appendix~\ref{app:test_time_ablation}).

\paragraph{Cross-dataset transfer.}
Cross-dataset gains are disproportionately large (+94\% H$\to$K vs.\ +29\% in-distribution) because in-distribution models can exploit dataset-specific $p(a)$ as a spurious cue; cross-dataset evaluation strips this crutch. The distilled FlanT5 canonicalizer matches or exceeds the LLM on both datasets (Table~\ref{tab:indist_h3d}), making MoCHA deployable without LLM infrastructure. We expect the $(s, a)$ decomposition to extend to other contrastive tasks with annotator-dependent captions.

\paragraph{Limitations.}
Canonicalization cannot compensate for insufficient motion coverage (K$\to$H remains $\sim$2--3\%), and the $s$/$a$ boundary is not always sharp---some annotations encode partially recoverable biomechanical cues (e.g., ``elderly gait'') that fall outside the current prompt specification. Jointly optimizing $C$ with the retrieval loss is a promising direction.

%==============================================================================
\section{Conclusion}
\label{sec:conclusion}
%==============================================================================

We have shown that caption supervision in motion-text retrieval is not deterministic but distributional: each caption mixes motion-recoverable content with annotator-specific style and inferred context, inducing within-motion embedding variance that weakens contrastive alignment. MoCHA reduces this variance by projecting captions onto their motion-recoverable content, producing tighter positive clusters and better-separated embeddings without modifying the retrieval architecture. Canonicalization is a general principle---even rule-based methods improve transfer---though learned canonicalizers (LLM and distilled FlanT5) provide the largest gains, achieving state-of-the-art results on HumanML3D, KIT-ML, and cross-dataset H$\to$K retrieval (+94\%).

%==============================================================================
% Appendices (included for cross-referencing)
%==============================================================================

%==============================================================================
% References
%==============================================================================
\clearpage
\bibliographystyle{splncs04}

\clearpage
%Same content as appendices, without preamble
\appendix

\section*{Appendices}

\section{Our Baseline Ablations}
\label{app:baseline_ablations}

Baseline ablations (architecture, temperature, self-similarity threshold) are consolidated into the main paper (Tables 3 and 6) for readability.

%----------------------------------------------------------------------
\section{LLM Ceiling: Best-Case with LLM at Train and Test}
\label{app:llm_ceiling}
%----------------------------------------------------------------------

The LLM ceiling (LLM canonicalization at both train and test) is reported as MoCHA (LLM) in the main paper's in-distribution results (Table 3).

%----------------------------------------------------------------------
\section{Evaluation Protocols}
\label{app:protocols}
%----------------------------------------------------------------------

\subsection{Protocol Definitions}
\label{app:protocol_definitions}

We evaluate retrieval under three protocols of increasing permissiveness:
\begin{enumerate}[nosep,leftmargin=*]
    \item \textbf{Full (1:1 diagonal)}: Each of $N$ motions is paired with exactly one caption. The retrieval task is to find the correct match among all $N$ candidates. This is the strictest protocol.
    \item \textbf{DsPair (dataset-pair)}: Motions with multiple captions are grouped; a match to \emph{any} caption in the group counts as correct. This reflects the multi-caption nature of HumanML3D~\cite{guo2022humanml3d} (avg.~3 captions/motion) and KIT-ML~\cite{plappert2016kit}.
    \item \textbf{Threshold}: Following TMR++~\cite{bensabath2024crossdataset}, we compute pairwise text similarity using \texttt{all-mpnet-base-v2}~\cite{reimers2019sentencebert} and group motions whose captions have cosine similarity $\geq 0.95$ (mapped to $\geq 0.90$ in raw cosine space). A match to any motion in the group counts as correct. This is the most permissive protocol, accounting for near-duplicate captions across different motions.
\end{enumerate}

%----------------------------------------------------------------------
\section{LLM Canonicalization Prompt}
\label{app:llm_prompt}
%----------------------------------------------------------------------

We use GPT-5.2 with temperature 0 and structured JSON output. The full prompt:

\begin{quote}\small\ttfamily
Convert each motion caption to a canonical form that preserves key motion details.\\[4pt]
KEEP (important for motion understanding):\\
- Action verbs: walk, step, throw, pick up, place, wipe, wave, punch, kick, dance, waltz, etc.\\
- Directions: forward, backward, left, right, up, down\\
- Limbs/body parts: right hand, left arm, both hands, right foot\\
- Objects being interacted with: cloth, item, ball, counter\\
- Poses: defensive pose, ready stance, crouch\\
- Repetition: twice, three times, repeatedly\\
- Manner when motion-relevant: quickly dodge, slow walk\\[4pt]
REMOVE (filler only):\\
- Subject words: person, man, woman, figure, someone, a, the\\
- Hedge phrases: seems to, appears to, looks like\\
- Unnecessary discourse: then, and then, after that\\[4pt]
Format: "verb [object] [limb] [direction] $\to$ next action..."\\[4pt]
EXAMPLES:\\
- "a person walks forward, then raises its right arm up and down twice" $\to$ "walk forward $\to$ raise right arm up-down twice"\\
- "person uses right hand to throw an item" $\to$ "throw item right hand"\\
- "a person picks up a cloth with the right hand, item with the left, then wipes it" $\to$ "pick up cloth right hand $\to$ pick up item left hand $\to$ wipe"
\end{quote}

For BABEL reverse expansion, see Appendix~\ref{app:reverse_prompt}.

%----------------------------------------------------------------------
\section{Reverse Expansion Prompt}
\label{app:reverse_prompt}
%----------------------------------------------------------------------

\begin{quote}\small\ttfamily
You are expanding short motion action labels into descriptive canonical motion captions.\\[4pt]
These action labels come from the BABEL motion dataset and are very short (1--3 words). You need to expand them into the canonical form used by a motion retrieval system.\\[4pt]
TARGET STYLE EXAMPLES (canonical motion captions from training data):\\
\{50 randomly sampled (original, canonical) pairs\}\\[4pt]
RULES:\\
1. Expand atomic labels into descriptive canonical forms matching the style above\\
2. Use the arrow notation for multi-step actions: "action1 -> action2"\\
3. Add plausible spatial details when naturally implied by the action\\
4. Keep it concise --- add only what's naturally implied\\[4pt]
Common expansions: "walk" $\to$ "walk forward", "stand" $\to$ "stand in place", "t pose" $\to$ "stand with arms extended horizontally", "transition" $\to$ "transition between poses"
\end{quote}

%----------------------------------------------------------------------
\section{Implementation Details}
\label{app:implementation}
%----------------------------------------------------------------------

\begin{table}[h]
\centering
\small
\caption{FlanT5~\cite{chung2022flant5} fine-tuning hyperparameters.}
\begin{tabular}{lcc}
\toprule
\textbf{Hyperparameter} & \textbf{FlanT5-PPT} & \textbf{FlanT5-Rev} \\
\midrule
Base model & FlanT5-base (250M) & FlanT5-PPT ckpt \\
Training pairs & 168K & $\sim$4.8K \\
Frozen layers & None & All exc.\ last 2 dec. \\
Learning rate & $3 \times 10^{-4}$ & $1 \times 10^{-5}$ \\
Batch size & 128 & 32 \\
Max epochs & 10 & 5 \\
LR schedule & Warmup + cosine & Warmup + cosine \\
Early stopping & Patience 3 & Patience 3 \\
Loss & Cross-entropy & Cross-entropy \\
\bottomrule
\end{tabular}
\end{table}

\begin{table}[h]
\centering
\small
\caption{MotionPatches (MoPa)~\cite{motionpatches} retrieval training hyperparameters.}
\begin{tabular}{lc}
\toprule
\textbf{Hyperparameter} & \textbf{Value} \\
\midrule
Motion encoder & ViT-B/16, 22 joints \\
Text encoder & DistilBERT (CLS pool) \\
Embedding dim & 256 \\
Loss & Symmetric InfoNCE \\
Temperature $\tau$ & 0.07 (frozen) \\
Self-sim threshold & 0.80 (mpnet) \\
Batch size & 128 \\
Learning rate & $10^{-5}$ \\
Motion LR & $10^{-6}$ (H3D) / $10^{-5}$ (KIT) \\
LR schedule & Cosine (per-batch) \\
Epochs & 50 \\
Gradient clipping & None \\
\bottomrule
\end{tabular}
\end{table}

\section{Additional Ablations and Analysis}
\label{app:ablations}

\subsection{Canonicalization Strategy Comparison}
\label{app:strategy_comparison}

\begin{table}[h]
\centering
\caption{\textbf{MoCHA (T5) canonicalization strategy comparison} (DsPair T2M R@1).}
\label{tab:ablation_app}
\small
\begin{tabular}{l cc}
\toprule
\textbf{Strategy} & \textbf{H3D} & \textbf{KIT} \\
\midrule
MoPa Baseline (orig text) & 10.80 & 14.02 \\
\hdashline
MoCHA Cardinal    & 12.66 & \best{23.66} \\
MoCHA Blend   & \best{13.30} & \second{22.14} \\
MoCHA Blend-Rev & 12.14 & 20.74 \\
\bottomrule
\end{tabular}
\end{table}

MoCHA Blend achieves the best balance of gains across both datasets, leading on H3D (13.30\%) while remaining competitive on KIT-ML (22.14\%). Cardinal leads on KIT-ML (23.66\%) but underperforms Blend on H3D. Main Table~3 reports MoCHA Blend as MoCHA (T5).

\subsection{FlanT5 Training Data: PPT}
\label{app:ppt_vs_para}

All FlanT5 canonicalizers in this work use the \textbf{PPT} (Paraphrases Plus Train) training regime: 168K pairs consisting of LLM paraphrases from training-split samples augmented with 38K original training captions paired with their LLM canonicals. The paraphrases are sourced from the TMR++~\cite{bensabath2024crossdataset} augmented caption data; the canonical target texts are our contribution, generated by our LLM canonicalization prompt (Appendix~\ref{app:llm_prompt}). Including original captions ensures the model learns the target mapping directly, rather than only seeing synthetic paraphrases.

\subsection{Complementarity with TMR Architecture}
\label{app:tmr}

\begin{table}[h]
\centering
\caption{\textbf{Canonicalization on TMR} (DsPair T2M R@1, canonical text at test). Evaluated at best validation epoch. $^\dagger$KIT baseline corrected after fixing case-sensitivity in mirror-caption grouping.}
\label{tab:tmr_app}
\small
\begin{tabular}{l cc}
\toprule
\textbf{Model} & \textbf{H3D} & \textbf{KIT} \\
\midrule
TMR Baseline (orig)      & 8.96  & 11.70$^\dagger$ \\
TMR Pure Cardinal        & 10.31 & \best{19.08} \\
TMR Blend                & \best{13.64} & 17.05 \\
TMR Blend-Rev            & 11.09 & \second{17.68} \\
\bottomrule
\end{tabular}
\end{table}

Canonicalization is architecture-agnostic. TMR~\cite{petrovich2023tmr} Blend improves H3D DsPair T2M R@1 from 8.96\% to 13.64\% (+52\%). On KIT-ML, TMR Pure Cardinal achieves 19.08\% from a corrected baseline of 11.70\% (+63\%). The gains are consistent despite TMR's fundamental architectural differences from MoPa: TMR uses 263-dim engineered features with a learned temperature and VAE regularization, whereas MoPa uses raw 3D joints with a frozen temperature.

\paragraph{Strategy-dependent patterns.} TMR's blend variant is the best strategy on H3D (13.64\%) and competitive on KIT-ML (17.05\%), while pure cardinal dominates KIT-ML (19.08\%) but lags on H3D (10.31\%). Blend-Rev underperforms relative to blend on both datasets (11.09 vs.\ 13.64 on H3D, 17.68 vs.\ 17.05 on KIT---effectively tied), suggesting blend training offers the most consistent gains across architectures.

\subsection{Training Stability Across Seeds}
\label{app:seed_variance}

\begin{table}[h]
\centering
\caption{\textbf{Retrieval performance across 3 random seeds} (DsPair Avg R@1). Each model evaluated at its own best validation epoch. ``+canon'' = canonical text at test time. $^\dagger$KIT baseline re-evaluated after correcting case-sensitivity in mirror-caption grouping.}
\label{tab:seed_variance}
\small
\begin{tabular}{l cccc}
\toprule
\textbf{Config} & \textbf{s42} & \textbf{s123} & \textbf{s456} & \textbf{Mean $\pm$ Std} \\
\midrule
\multicolumn{5}{l}{\textit{HumanML3D native}} \\
\quad Baseline (orig)         & 11.72 & 11.18 & 10.69 & 11.20 $\pm$ 0.52 \\
\quad MoCHA (LLM) Blend-Rev   & 14.09 & 11.72 & 13.14 & 12.98 $\pm$ 1.19 \\
\quad MoCHA (T5) Blend-Rev    & 13.73 & 12.24 & 12.20 & \best{12.72 $\pm$ 0.87} \\
\midrule
\multicolumn{5}{l}{\textit{KIT-ML native}} \\
\quad Baseline (orig)         & 14.31$^\dagger$  & 11.58$^\dagger$  & 13.87$^\dagger$  & 13.25 $\pm$ 1.20$^\dagger$ \\
\quad MoCHA (LLM) Blend-Rev   & 20.30 & 16.80 & 20.81 & \best{19.30 $\pm$ 2.18} \\
\quad MoCHA (T5) Blend-Rev    & 16.03 & 15.97 & 17.69 & 16.56 $\pm$ 0.98 \\
\bottomrule
\end{tabular}
\end{table}

Note: the main paper reports last-epoch (epoch 50) results for consistency; here we evaluate at each run's best validation epoch to isolate convergence-speed variance from final performance. This accounts for some difference in absolute numbers but enables a fair seed-to-seed comparison.

The retrieval gains are robust across seeds (Table~\ref{tab:seed_variance}). On KIT-ML, MoCHA (LLM) Blend-Rev achieves 19.30$\pm$2.18 vs.\ corrected baseline 13.25$\pm$1.20---a 1.5$\times$ improvement that holds across all 3 seeds. MoCHA (T5) Blend-Rev has tighter variance (16.56$\pm$0.98) despite similar mean gains. On H3D, gains are smaller but consistent: +1.8pp (LLM) and +1.5pp (T5) over the 11.20 baseline. Notably, the T5 variant achieves lower test-time variance than LLM on both datasets (0.87 vs.\ 1.19 on H3D, 0.98 vs.\ 2.18 on KIT), suggesting the distilled canonicalizer produces more consistent text normalization than the LLM.

\section{BABEL: Reverse Expansion and Cross-Dataset Results}
\label{app:crossds_babel}

\subsection{Reverse Expansion for Short-Caption Datasets}
\label{sec:reverse_expansion_method}

Short labels (\eg, BABEL's ``walk'') are already low-variance but under-specified.
\emph{Reverse expansion} maps a short label to a canonical-style description (\eg, ``walk''~$\to$~``walk forward'') via an LLM prompt, distilled into a seq2seq model for LLM-free inference. This is essential for bridging the domain gap to BABEL~\cite{punnakkal2021babel}, whose captions are ultra-short atomic labels fundamentally different from the verbose descriptions in HumanML3D and KIT-ML.

\subsection{BABEL Cross-Dataset Results}

Table~\ref{tab:crossds_full} shows cross-dataset transfer results involving BABEL. BABEL results should be interpreted with caution: of 23{,}732 test captions, only 4{,}833 are unique (86\% duplicates), inflating threshold metrics by 141$\times$ on H$\to$B and 168$\times$ on K$\to$B. Despite these caveats, relative comparisons between methods remain valid, and canonicalization with reverse expansion substantially improves BABEL transfer: H$\to$B from 15.84\% to 26.09\% (+65\%), K$\to$B from 12.76\% to 26.80\% (+110\%), and B$\to$B from 26.15\% to 39.31\% (+50\%).

\begin{table}[t]
\centering
\caption{\textbf{BABEL cross-dataset retrieval} (Threshold T2M protocol per TMR++). X$\to$B comparisons use last epoch as BABEL R@1 is unstable due to caption duplicity. BABEL numbers are inflated by 86\% caption duplication (see text). H$\to$K and K$\to$H results without BABEL are reported in the main paper (Table~4).}
\label{tab:crossds_full}
\small
\setlength{\tabcolsep}{3pt}
\begin{tabular}{ll l ccc}
\toprule
& & & \multicolumn{3}{c}{\textbf{T2M}} \\
\cmidrule(lr){4-6}
\textbf{Train} & \textbf{Model} & \textbf{Test} & R@1 & R@5 & R@10 \\
\midrule
\multirow{3}{*}{H3D}
& MoPa                  & BABEL & 15.84 & \second{43.32} & \second{49.31} \\
& MoCHA (T5 blend-rev) & BABEL & \second{22.36} & 35.57 & 42.07 \\
& MoCHA (LLM blend) & BABEL & \best{26.09} & \best{46.76} & \best{55.92} \\
\midrule
\multirow{3}{*}{KIT}
& MoPa                  & BABEL & 12.76 & \second{29.93} & \second{36.99} \\
& MoCHA (T5 blend-rev)  & BABEL & \second{25.11} & 29.66 & 30.90 \\
& MoCHA (LLM blend-rev) & BABEL & \best{26.80} & \best{31.72} & \best{46.48} \\
\midrule
\multirow{3}{*}{BABEL}
& MoPa                  & H3D   & \second{3.40} & 7.71 & 12.14 \\
& MoCHA (T5) & H3D   & 3.03 & \second{10.01} & \second{16.29} \\
& MoCHA (LLM) & H3D   & \best{3.70} & \best{10.95} & \best{16.61} \\
\hdashline
& MoPa                  & KIT   & \second{10.31} & 28.24 & 37.91 \\
& MoCHA (T5) & KIT   & \best{13.49} & \best{30.15} & \best{41.22} \\
& MoCHA (LLM) & KIT   & 9.41 & \second{28.37} & \second{40.20} \\
\hdashline
& MoPa                  & BABEL & \second{26.15} & 44.29 & 57.34 \\
& MoCHA (T5) & BABEL & \best{39.31} & \second{46.77} & \second{58.18} \\
& MoCHA (LLM) & BABEL & 24.25 & \best{53.74} & \best{58.23} \\
\bottomrule
\end{tabular}
\end{table}

\section{Test-Time Caption Ablation}
\label{app:test_time_ablation}

\begin{table}[t]
\centering
\caption{\textbf{Test-time text mode ablation} (DsPair T2M R@1, epoch 50). Each row is a different training strategy (LLM-trained models); columns show performance under each test-time text mode. Note: Main Table~3 MoCHA (T5) reports a separately-trained FlanT5-PPT Blend model (13.30\%/22.14\%); see Appendix~\ref{app:strategy_comparison} for all FlanT5-PPT variants.}
\label{tab:test_text}
\small
\begin{tabular}{l ccc ccc}
\toprule
& \multicolumn{3}{c}{\textbf{H3D}} & \multicolumn{3}{c}{\textbf{KIT}} \\
\cmidrule(lr){2-4} \cmidrule(lr){5-7}
\textbf{Train} & Orig & LLM & T5 & Orig & LLM & T5 \\
\midrule
MoPa Baseline     & 11.13 & 7.30 & 9.24 & 8.91 & 15.39 & 13.87 \\
MoCHA Cardinal    & 6.16 & 12.52 & 11.93 & 6.36 & 23.92 & 23.03 \\
MoCHA Blend       & 11.82 & 13.21 & 12.18 & 10.69 & 15.39 & 18.32 \\
MoCHA Blend-Rev   & 11.45 & 13.91 & \best{14.03} & 8.65 & 24.30 & \best{25.83} \\
\bottomrule
\end{tabular}
\end{table}

Table~\ref{tab:test_text} reveals how each training strategy responds to different test-time text. The baseline degrades with canonical text on H3D (11.13$\to$7.30 with LLM) because the model was trained on verbose captions. MoCHA Cardinal collapses on original text (6.16 on H3D, 6.36 on KIT), motivating blend training. MoCHA Blend-Rev is the most robust: it maintains reasonable performance on original text while excelling with canonical text (14.03 / 25.83). Interestingly, T5 sometimes exceeds LLM at test time despite being a distilled approximation.

\section{Full Embedding Space Geometry}
\label{app:embedding_geometry_full}

Table~\ref{tab:embedding_geometry_full} shows the embedding space geometry for baseline and MoCHA Blend-Rev models, each evaluated with both original and canonical captions. MoCHA Blend-Rev surpasses the baseline on all metrics regardless of test-time text mode, confirming that the improved embedding structure comes from training, not from canonicalization at test time alone.

\begin{table}[t]
\centering
\caption{\textbf{Full embedding space geometry} in the trained encoder's 256-dim retrieval space. Intra: text-text similarity within same-motion captions. Align: text-motion cosine similarity. Inter NN: nearest negative similarity. Sep: Intra / Inter NN (higher = better separation).}
\label{tab:embedding_geometry_full}
\small
\begin{tabular}{ll cccc}
\toprule
& \textbf{Condition} & Intra & Align & Inter NN & Sep \\
\midrule
\multicolumn{6}{l}{\textit{HumanML3D (4{,}382 multi-caption motions)}} \\
\quad & Baseline + orig       & 0.413 & 0.601 & 0.976 & 0.423 \\
\quad & Blend-Rev + orig      & \second{0.444} & \second{0.634} & 0.970 & \best{0.458} \\
\quad & Blend-Rev + canon     & \best{0.444} & \best{0.636} & 0.971 & \second{0.457} \\
\midrule
\multicolumn{6}{l}{\textit{KIT-ML (393 multi-caption motions)}} \\
\quad & Baseline + orig       & 0.456 & 0.515 & 0.941 & 0.484 \\
\quad & Blend-Rev + orig      & \second{0.557} & \best{0.553} & 0.937 & \second{0.595} \\
\quad & Blend-Rev + canon     & \best{0.566} & \best{0.553} & 0.938 & \best{0.604} \\
\bottomrule
\end{tabular}
\end{table}

\section{Cross-Dataset Caption Alignment}
\label{app:crossds_caption_alignment}

Using AMASS path mapping, we identify 5{,}766 motions present in both HumanML3D and KIT-ML. For each matched pair, we embed captions from both datasets through the baseline text encoder and measure cross-dataset cosine similarity. Canonicalization increases mean cross-dataset similarity by +12.5\% (FlanT5: 0.444 $\to$ 0.500) and +14.3\% (LLM: 0.444 $\to$ 0.508). The same physical motion, described as ``a person nervously walks forward'' in H3D and ``walk forward'' in KIT, becomes measurably more similar after denoising. This is the geometric mechanism behind the cross-dataset transfer gains.

\section{Training Dynamics and Overfitting}
\label{app:training_dynamics}

Training dynamics and convergence analysis are incorporated into the main paper (Section 4) to support the narrative flow from variance reduction to retrieval gains.

\section{BABEL Transfer Ceiling Analysis}
\label{app:babel_ceiling}

The BABEL transfer ceiling analysis (LLM vs.\ T5 reverse expansion) is incorporated into the cross-dataset results in Appendix~\ref{app:crossds_babel}.

\section{Linguistic Properties of Canonicalization}
\label{app:canon_properties}

We analyze linguistic properties of the canonicalization mapping to understand what changes drive the retrieval improvements.

\setcounter{subsection}{2}
\subsection{Caption Length vs.\ Retrieval Gain}
\label{app:caption_length}

A natural hypothesis is that canonicalization helps primarily by shortening verbose captions, reducing noise in long descriptions. We test this by partitioning test queries into three length bins based on original caption word count (short: 1--6 words, medium: 7--12, long: 13+), then comparing per-bin R@1 between the baseline and blend-rev models. We extract text and motion embeddings from both models, compute per-query DsPair ranks (matching to any motion sharing the same caption), and measure the R@1 improvement ($\Delta$) for each bin. We also compute the Pearson correlation between per-query caption length and per-query R@1 change (binary: hit or miss).

\begin{table}[!ht]
\centering
\caption{\textbf{Caption length vs.\ R@1 gain} (DsPair T2M R@1). $\Delta$: Blend-Rev R@1 minus baseline R@1 (percentage points).}
\label{tab:caption_length}
\small
\begin{tabular}{ll ccc c}
\toprule
\textbf{Data} & \textbf{Bin} & N & Base & Blend-Rev & $\Delta$ \\
\midrule
H3D & short (1--6) & 1{,}145 & 17.78 & 17.71 & $-$0.07 \\
H3D & medium (7--12) & 2{,}382 & 12.97 & 16.73 & +3.76 \\
H3D & long (13+) & 856 & 12.07 & 12.68 & +0.61 \\
\midrule
KIT & short (1--6) & 523 & 13.21 & 21.65 & +8.44 \\
KIT & medium (7--12) & 161 & 19.52 & 32.11 & +12.59 \\
KIT & long (13+) & 102 & 21.22 & 17.65 & $-$3.57 \\
\bottomrule
\end{tabular}
\end{table}

Table~\ref{tab:caption_length} shows that gains are \emph{not} driven by caption length. Medium-length captions (7--12 words) benefit most on both datasets (H3D: +3.76pp, KIT: +12.59pp), while short captions ($\leq$6 words) show negligible change on H3D ($-$0.07pp) despite being compressed the most in relative terms. The Pearson correlation between caption length and R@1 gain is weak and non-significant on both datasets (H3D: $r$=0.023, $p$=0.14; KIT: $r$=$-$0.052, n.s.). If canonicalization worked primarily through compression, we would expect the longest captions---which lose the most words---to show the largest gains. Instead, the pattern suggests that canonicalization helps through \emph{semantic normalization}: standardizing \emph{how} a motion is described matters more than \emph{how many words} are used.

\subsection{Gradient Concentration under InfoNCE}
\label{app:gradient_concentration}

We measure two properties of the trained models' InfoNCE landscape on the test set: \textbf{softmax entropy} (lower = more concentrated probability mass, more gradient signal toward the correct positive) and \textbf{P(positive)} (the softmax probability assigned to the correct match).

\begin{table}[!ht]
\centering
\caption{\textbf{InfoNCE gradient concentration.} Softmax entropy and P(positive) measured on trained models' test-set similarity matrices. Lower entropy and higher P(positive) indicate more concentrated gradient signal toward the correct positive.}
\label{tab:gradient_concentration}
\small
\begin{tabular}{l cc cc}
\toprule
& \multicolumn{2}{c}{\textbf{HumanML3D}} & \multicolumn{2}{c}{\textbf{KIT-ML}} \\
\cmidrule(lr){2-3} \cmidrule(lr){4-5}
\textbf{Model} & Entropy & P(+) & Entropy & P(+) \\
\midrule
MoPa             & 6.285 & 0.0158 & 4.939 & 0.0300 \\
$+$Paraphrases   & 6.093 & 0.0156 & 4.901 & 0.0288 \\
MoCHA (blend-rev) & \best{6.034} & \best{0.0190} & \best{4.673} & \best{0.0337} \\
\bottomrule
\end{tabular}
\end{table}

Table~\ref{tab:gradient_concentration} confirms the theoretical prediction. MoCHA (blend-rev) achieves the lowest softmax entropy on both datasets (6.034 vs.\ 6.285 baseline, 6.093 augmented on H3D) and the highest P(positive) (0.0190 vs.\ 0.0158 baseline, 0.0156 augmented on H3D---a 21\% relative increase). Paraphrase augmentation barely changes the gradient concentration relative to the baseline (and actually \emph{decreases} P(positive) on KIT-ML: 0.0288 vs.\ 0.0300), while canonicalization substantially sharpens the InfoNCE distribution. This provides a gradient-level explanation for the retrieval gains: canonicalization concentrates more learning signal on the correct motion-text correspondence, while augmentation dilutes it across paraphrase variants.

\section{Failure Mode Analysis}
\label{app:failure_modes}

The canonicalization operator $C$ is a many-to-one mapping and cannot be perfect: some captions lose motion-relevant detail in the projection from $t$ to $C(t)$. We analyze failure modes by categorizing the content stripped during canonicalization across all 4{,}377 modified HumanML3D test captions.

The most frequently removed words are function words and generic subjects (``a,'' ``person,'' ``the,'' ``and'')---content with no motion-discriminative value. Among content words, 503 directional terms, 310 body part references, and 203 manner adverbs are stripped. Most of these removals are redundant (e.g., ``a person'' removed when the action verb already implies an agent), but some represent genuine information loss:

\begin{itemize}[nosep,leftmargin=*]
\item \textbf{Over-compression}: ``a person with both feet on the ground with both knees bended'' $\to$ ``move both feet side to side'' loses the static pose semantics entirely.
\item \textbf{Manner loss}: ``person is working on their boxing form'' $\to$ ``boxing stance'' drops the iterative practice implied by ``working on.''
\end{itemize}

However, these cases are outweighed by beneficial denoising:
\begin{itemize}[nosep,leftmargin=*]
\item ``a person walks forward, slowly.'' $\to$ ``walk forward slow'' (strips filler, preserves semantics).
\item ``a person doges to the left, then doges to the right.'' $\to$ ``dodge left $\to$ dodge right'' (corrects misspelling, standardizes format).
\end{itemize}

The net effect is positive across all recall ranks: the operator strips more noise than signal. This is consistent with the supervision noise model---the majority of caption content that varies across annotators is non-kinematic $(a)$, so a denoising operator with imperfect precision still yields net improvement.

\end{document}